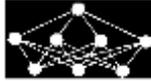

# AN ACCURACY-ENHANCED STEMMING ALGORITHM FOR ARABIC INFORMATION RETRIEVAL

*Sadik Bessou\*, Mohamed Touahria\**

**Abstract:** This paper provides a method for indexing and retrieving Arabic texts, based on natural language processing. Our approach exploits the notion of template in word stemming and replaces the words by their stems. This technique has proven to be effective, since it has returned significant relevant retrieval results by decreasing silence during the retrieval phase. Series of experiments have been conducted to test the performance of the proposed algorithm ESAIR (Enhanced Stemmer for Arabic Information Retrieval). The results obtained indicate that the algorithm extracts the exact root with an accuracy rate up to 96% and hence, improving information retrieval.

## 1. Introduction

The accessible electronic documents in websites constitute a field of documentary research that is extensively growing [1]. According to the web father, these documents are intended to be decoded by humans rather than being data that can be automatically analysed [2]. The challenge is to automatically extract the information contained in these documents which are written in natural language, since "the power of the natural language creates an obstacle to its use for data processing" [3]. Nowadays, various languages are successfully processed. However, indexing Arabic language documents remains a big challenge towards its integration in the information technology, given its power and its wealth.

Automatic indexing of Arabic documents raises major problems [4, 5]:
- the problem of ambiguity caused by the absence of vowels, which requires complex morphological rules [6]; and
- the problem of inflected forms recognition, because Arabic derivational morphology is productive [7].

---

\* Sadik Bessou, Mohamed Touahria
Department of Computer Science, University of Setif1, SETIF, 19000, Algeria, E-mail: `bessou.s@gmail.com, touahria_momo@yahoo.fr`

The objectives for this treatment are:
- the intelligent recognition of various word forms, without using a dictionary of inflected forms, i.e. without knowing these words in advance;
- the reduction of ambiguity as much as possible;
- the identification of the information contained in any text and its representation by means of indexes, in a restricted space with regard to the space where these documents are stored [8];
- the reduction of response time.

Information retrieval systems represent, store, and organize documents in such a way that they can localise a set of relevant documents satisfying the needs of a user as expressed in a query [9, 10, 11].

## 2. Analysis Approach

To fully understand a text, it is required to perform a complete linguistic analysis. The linguistic techniques allow for a better understanding of the indexed documents. However, such methods require powerful natural language processing algorithms and considerable processing time.

The suitable technique used for information retrieval and machine translation is stemming. This technique consists of removing the last characters of words (considered as describing word flexions). Some stemmers use complete morphological knowledge (suffixes, prefixes). Instead of indexing a text by words, we can index it by the corresponding stems.

The main difficulty of a retrieval system is to establish a correspondence between the information requested by a user and that contained in documents. To accomplish this, the generally used method is matching words in the query with those representing the contents of documents as shown in *Fig. 1*. Considering this mechanism, based on a simple comparison of strings, information retrieval systems confront the problem linked to the complexity of the natural language. It concerns the possibility offered by the natural language to supply various expressions with the same concept. A relevant document can contain terms which differ from those of the query, though semantically close. A solution to tackle this problem is to turn to natural language processing.

The aim of introducing linguistic knowledge in information retrieval is to enter more robust and more pertinent descriptors than simple strings [12].

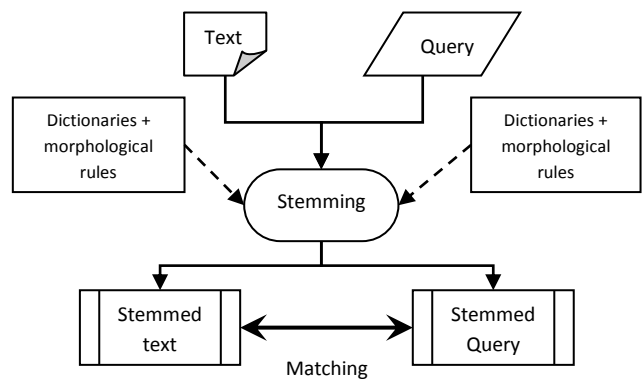

Fig. 1 *Words stemming and query matching.*

Our algorithm ESAIR (Enhanced Stemmer for Arabic Information Retrieval) decomposes words into morphemes [13, 14], without necessarily considering grammatical links between them. It proceeds as follows:

- normalization step: it transforms the document into a format that is more easily tractable [15]. This stage is a delicate step, due to the fact that Arabic is an inflected language and is productive;

- lexical analysis step: it allows verifying (a) if an item belongs to the language, and (b) the compatibility between the various constituents of the word. The morphological analyser works with the help of lexical dictionaries [dictionaries of roots (triliterals), templates (the number of patterns in Arabic is finite), stop words, grammatical words, and specific words];

- indexing step: it generates the indexes by grouping the words together based on their stems.

To verify whether a word belongs to Arabic lexemes (with the exception of the proper nouns, some common nouns, and grammatical words), we need to extract its root and the corresponding template. According to this method, the proposed work is based on three main steps: cutting, search for templates and roots, and index generation.

## 2.1 Cutting

To resolve the ambiguity, (Aljlayl and Frieder) show that light stemming (an approach based on the extraction of prefixes and suffixes) significantly outperforms approaches based on root detection in the field of information retrieval [16].

Cutting a word requires the extraction of its various parts (prefix, root, and suffix). The principles of cutting are as follows:

Cutting the **word** in (proclitic+ base 1+ enclitic) entails locating all the proclitics and enclitics that appear in the word.

**Base 1** is generally a root containing prefixes, infixes, and suffixes that could be systematized as follows: (prefix+ base 2+ suffix);

**Base 2** could be cut into a root and a template, i.e. finding out a pattern amongst the entries stored in the dictionary of templates.

**Example** The word: "أستخرجانها" [ASSATOKHRIJANIHA] could be decomposed into "ها"+"تخرجان"+"أسـ", where the proclitic= "أسـ", base 1 ="تخرجان", and the enclitic="ها".

Furthermore, base 1: "تخرجان" could also be decomposed into "تـ"+"خرج"+"ان", where the prefix="تـ", base 2="خرج", and suffix="ان".

1) Recognition of proclitics and enclitics

a) Search for proclitics and enclitics

The list of proclitics and enclitics of the Arabic language is limited. We can use the list proposed by Darwish [17], some of which were used for stemming by Chen and Habash [18, 19]. Tab. 1 shows the list of proclitics and enclitics.

Tab. 1 *List of proclitics and enclitics.*

| Proclitics | ' ' | بـ | كـ | لـ | فـ | سـ | أ | الـ | كالـ | لل | فبـ | فسـ | فالـ |
|---|---|---|---|---|---|---|---|---|---|---|---|---|---|
|  |  |  |  |  |  | فكـ | فل | فلل | أف | أسـ | فبالـ | فكالـ | بالـ |
| Enclitics | ' ' | ه | ي | ك | هم | هن | هما | ها | كم | كن | كما | ني | نا |

Cutting a word into "proclitic+ base 1+ enclitic" is not limited to the search for a proclitic or an enclitic within a list. It also requires a verification of compatibility between proclitics and enclitics that appear in the word to retain only compatible couples.

b) Test of compatibility

After the extraction of the proclitic P and the enclitic E from the word under analysis, we combine these two sub-strings into one single string C (C= P+ E). In order to have a correct decomposition, this string must not appear in the table of incompatibilities [20]. If the string C appears in the table, this implies that the decomposition is erroneous.

The table of incompatibilities includes the incompatible proclitics and enclitics merged in one single string as mentioned in Tab. 2.

Tab. 2 *Sample of incompatibilities table.*

| String |
|---|
| بني |
| ………. |
| كالهما |
| ………. |
| فكالنا |

**Example**

The word "فكتابهما" [FAKITABOHOMA] is decomposed into the proclitic= "فـ" and the enclitic="هما". The combination of the proclitic and the enclitic generates the string C="فهما". This string does not appear in the table of incompatibilities and therefore the decomposition is accepted.

c) Principle of analysis

Relying on the word cutting and compatibility operations, we must cancel the decomposition if it is erroneous. If correct, it is stored. Other new decompositions will be processed, and compared with the stored ones. A reliable decomposition is then, obtained.

2) Recognition of prefixes and suffixes

The principle at this stage is technically the same as the previous one, except for the incompatibilities table which is of prefixes and suffixes.

## 2.2 Search for templates and roots

1) Search for the template

For a given word X, a template I corresponds to that word if:

- the length of the template is equal to that of the word, and
- all the letters corresponding to the infix positions in the dictionary of templates are in the same positions in the word X as illustrated in Fig. 2.

The example below elucidates how the process operates:

The Processing of the word = ' صالح ' [SALIH] requires the inspection of all the records which have the same length as that word until it encounters the template 'فاعل ' [FAIIL]. The corresponding field (infix positions) is '2'. The letter ' ا ' is in the position 2 in the word 'صالح '. Thus, it is practically the best template.

2) Search for the root

After determining the template, root extraction limits itself to the removal of all letters corresponding to infix positions in the word intended to be decomposed.

The word 'مفاتيح ' [MAFATIH] has as template ' مفاعيل ' [MAFAIIL]. The aforementioned template corresponds to the code '135' in the infix positions column as shown in Fig. 2.

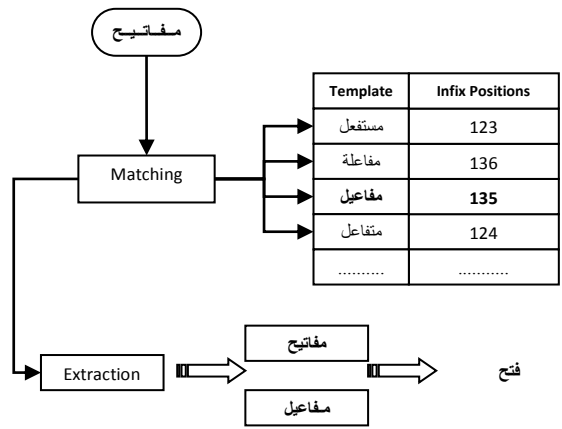

Fig. 2 *Template and root search.*

The Subtraction of the letter 'مـ ', from position 1 in the word 'مفاتيح ' [MAFATIH], and the letter 'ا ', from position 3, as well as the letter 'ـيـ ', from position 5, yields 'فتح ' [FTH]. Consequently, the correct root of the word 'مفاتيح ' [MAFATIH], which is the root 'فتح ' [FTH] is attained.

It is worth mentioning that there are more steps, which can be introduced if necessary: treatment of compatibility between proclitics / prefixes, enclitics / suffixes, and disambiguation in case there are various interpretations.

## 2.3 Index generation

Descriptors can be text words, stems, concepts and, more rarely, N-grams, or still contexts (the case of "Latent Semantic Indexing" or methods based on Correspondence Analysis). The models using words can fit the English language (few flexions, few homographs), but they turn out to be largely insufficient in other languages (particularly for agglutinative languages as Arabic). We can then use the stems for better performances. An advanced linguistic analysis is necessary to resolve certain ambiguities [21].

The aim of the morphological level is to reduce the number of analysed words by stemming, and to remove stop words to lighten the index, and to reduce silence (formula 1) in information retrieval [22].

$$Silence = \frac{non\ retrieved\ relevant\ documents}{relevant\ documents} \qquad (1)$$

## 3. Illustration

In this section, the results obtained through the adopted approach are discussed in detail: ESAIR analyses and converts any query into elements of the indexing language whenever a user formulates a request. ESAIR compares the query elements with those of the compiled documents, determines the degree of similarity, and then selects the elements that have the highest degree.

The advantage of ESAIR is the decrease of silence during retrieval. Even words not mentioned in a query can appear in retrieval results.

Consider the following text with 162 words:

(مما صح عنه عليه الصلاة و السلام قوله: " عليكم بالصدق، فان الصدق يهدي إلى البر. و إن البر يهدي إلى التقوى. و لا يزال الرجل يصدق و يتحرى الصدق، حتى يكتب عند الله صديقا. و إياكم والكذب، فان الكذب يهدي إلى الفجور. و إن الفجور يهدي إلى النار. و لا يزال الرجل يكذب و يتحرى الكذب، حتى يكتب عند الله كذابا ". في هذا الحديث الشريف، دعوة قوية من النبي صلى الله عليه و سلم إلى إتباع الصدق و التحلي به قولا و فعلا. بل على الرجل الفطن الكيس أن يفتش و ينقب و يتحرى أثاره في كل زمان و مكان. لأن ذلك يهديه و يقوده إلى البر و الخير، فان فعل ذلك كان مع الصديقين و الشهداء و حسن أولئك رفيقا أي جوارا و صحبة في الجنة. و من معاني الصدق كذلك التصديق بكل ما أمر الله تعالى الإيمان به. و من معانيه كذلك الصدقة و الإنفاق في سبل و أوجه الخير الكثيرة، مصداقا لقوله تعالى " إنما الصدقات للفقراء و المساكين ".)

Tab. 3 shows the text after stop words removal. It contains 72 words that represent 44% of the original text.

Tab. 3 *Text after stop words removal.*

| بالصدق | يصدق | يهدي | الكذب | الصدق | أثاره | الصديقين | التصديق | الخير |
|---|---|---|---|---|---|---|---|---|
| الصدق | يتحرى | الفجور | يكتب | التحلي | زمان | الشهداء | الله | الكثيرة |
| يهدي | الصدق | الفجور | الله | قولا | مكان | حسن | الإيمان | الصدقات |
| البر | يكتب | يهدي | كذابا | فعلا | يهديه | رفيقا | معانيه | للفقراء |
| البر | الله | النار | دعوة | الرجل | يقوده | جوارا | الصدقة | المساكين |
| يهدي | صديقا | الرجل | النبي | يفتش | البر | صحبة | الإنفاق | قوية |
| التقوى | الكذب | يكذب | الله | ينقب | الخير | الجنة | أوجه | الفطن |
| الرجل | الكذب | يتحرى | إتباع | يتحرى | فعل | الصدق | سبل | الكيس |

Tab. 4 shows the text after applying the previous treatment steps on each word. The second text contains 43 words that represent 26% of the original text.

Tab. 4 *Text after applying ESAIR.*

| صدق | رجل | صديق | جنة | حلي | أثار | شهداء | إنفاق | فطن |
|---|---|---|---|---|---|---|---|---|
| هدي | حرى | كذب | إيمان | قول | زمان | حسن | أوجه | كيس |
| بر | كتب | فجور | دعو | فعل | مكان | رفيق | سبل | كثير |
| تقوى | اله | نار | نبي | فتش | قاد | جوار | فقراء | |
| قوي | تصديق | كذاب | إتباع | نقب | خير | صحب | مساكين | |

In the following example, we wanted to include as many words as possible that have a same root and treat it in comparison with another one with completely different words.

The index size decrease is proportional to the text. If the text contains a variety of terms derived from the same root, as in Text 1:

Text 1: كتب الكاتب في مكتبه بالكاتبة كل الكتب غير المكتوبة في المكتبة [KATABA ALKATIBO FI MAKTABIHI BILKATIBATI KOULLA ALKOUTOUBI GHAIRA ALMAKTOUBATI FI ALMAKTABATI].

The index 1 evolves as follows: Index 1: كتب [KTB], and the reduction leans towards 98%, as shown in Fig. 3.

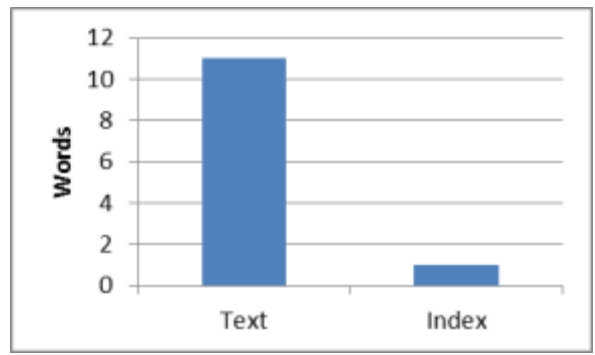

Fig. 3 *Text 1 vs. Index 1.*

If the text does not contain two words with a common root or base, as in text 2:

Text 2:تقدم الدولة الجزائرية حاليا تحفيزات كبيرة للنهوض بالبحث العلمي في الجزائر [TOKADIMO ADAWLATO ALJAZIRYATO HALIEN TAHFIZET KABIRA LINOHODI BILBAHTI ALILMI FI ALJAZAIR]

The index 2 evolves as follows: Index 2: [قدم، دول، حفز، نهض، بحث، علم، الجزائر] [KDM, DWL, HFZ, NHD, BHT, ALM, ALJAZAIR], and the reduction leans towards 0%, as shown in Fig. 4.

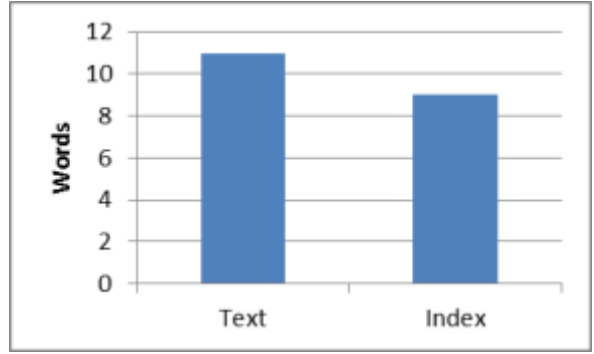

Fig. 4 *Text 2 vs. Index 2.*

The user's query is processed through all the indexing stages, including those of morphological analysis. Then, the query entries are indexed in a list that it is compared with document indexes as in the example:
Text 1: (يساهم المعلم بشكل مباشر في بناء الأجيال) [YOSSAHIMO ALMOALIMO BICHAKLIN MOBACHIR FI BINAI ALAJIALI]
Text 2: (أقام المدير حفلا على شرف المعلمات المتقاعدات) [AKAMA ALMODIRO HAFLEN ALA CHARAFI ALMOALIMATI ALMOTAKAIDATI]
Text 3: (زار التلاميذ و المعلمون مركزا للبحث العلمي) [ZARA ATALAMIDO WA ALMOALIMONA MARKAZEN LILBAHTI ALILMI]
Query: (معلم) [MOALIM]
If matching is verbatim, no text is found (0/3), contrariwise with ESAIR, the three texts (3/3) are obtained, since the stemming of words (المعلم ، المعلمات، المعلمون) [ALMOALIMO, ALMOALIMATO, ALMOALIMONA] gives the same word (معلم) [MOALIM].

Considering the previous text (Tab. 3), a set of 49 other texts, and the query = " الصدق في القول" [ASSIDKO FI ALKAOULI], we deduce: the number of relevant documents retrieved is 9 out of 12, the number of reported documents, given as an answer, is 14 documents. Therefore, Precision = (9/14) = 0.64, Recall = (9/12) = 0.75, and Silence = (3/12) = 0.25. The accuracy calculated on the previous text is: (69/72) = 0.958.

## 4. Experiments and Results

Experiments are performed by executing ESAIR on randomly selected documents of Essex Arabic Summaries Corpus (EASC). EASC is Arabic natural language resources developed at the University of Essex, United Kingdom. It contains 153 articles taken from "Alwatan" and "Alrai" newspapers, which covered different subjects: education, science and technology, finance, health, politics, religion, and sports. Each document contains an average of 389 words, with 59548 words in the corpus. We manually extracted word roots for purposes of comparison with ESAIR's results. A set of 25 queries, with their relevance judgments created to search for particular information, was used to evaluate the proposed method.
In Tab. 5, we provide a comparison measured in average precision and recall between ESAIR and no stemming method.

Tab. 5 *Average precision and recall.*

| Stemmer | Precision | Recall |
|---------|-----------|--------|
| ESAIR   | 0.5732    | 0.6916 |
| NoStem  | 0.4328    | 0.4152 |

The results clearly suggest that the proposed algorithm outperforms the no stemming method. This suggests that stemming has a substantial effect on information retrieval for highly inflected languages such as Arabic.
Fig. 5 shows precision at 11 recall points for ESAIR and no stemming method.

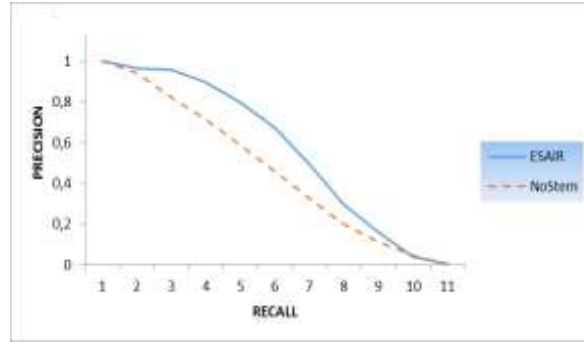

Fig. 5 *11 point precision for ESAIR and NoStem.*

After manually reviewing all resulting stems, accuracy for each document was calculated using the following formula:

$$\text{Accuracy} = \frac{\sum \text{correctly stemmed words}}{\text{total number of valid words}} \quad (2)$$

The results obtained indicate that the algorithm extracts the exact root with an accuracy rate up to 96% and hence, improving information retrieval.

## 5. Conclusion

This work tested a method based on the linguistic notion of template to information retrieval. It allowed us to index texts with small size indexes, representing the relevant information, and to analyse the query and match it with these indexes. This approach does not require a prior knowledge of words and can establish relations between words without the user's specification. It also, ensures substantial decrease of silence.
The algorithm was tested, using thousands of Arabic words taken from EASC corpus of 153 articles from "Alwatan" and "Alrai" newspapers. Human expert judgments were used to evaluate the results. The algorithm extracted the proper roots with a 96% accuracy. Stemming results in significant improvements in retrieval effectiveness of Arabic information retrieval systems with a precision= 0.5732. The difference in the performance between the proposed approach and the no stemming method is statistically significant.

### Acknowledgement

We would like to thank Dr Guy Tremblay from Université du Québec à Montréal (UQAM), Canada for his valuable support while performing experiments at LATECE laboratory, and Dr Mahmoud El haj from University of Essex, United Kingdom for providing us with the test corpus.